\documentclass[10pt,twocolumn,letterpaper]{article}

\usepackage{wacv}
\usepackage{times}
\usepackage{epsfig}
\usepackage{graphicx}
\usepackage{amsmath}
\usepackage{amssymb}
\usepackage{enumitem}
\usepackage{multirow}
\usepackage{algorithm}
\usepackage{algorithmic}
\usepackage{caption}
\usepackage{subcaption}
\usepackage{tabularx}
\usepackage[normalem]{ulem}
\usepackage[dvipsnames]{xcolor}

\usepackage{hyperref}
\hypersetup{
    colorlinks=true,
    linkcolor=blue,
    filecolor=magenta,      
    urlcolor=Rhodamine,
}
 
\wacvfinalcopy % *** Uncomment this line for the final submission

\def\wacvPaperID{508} % *** Enter the wacv Paper ID here

\ifwacvfinal\fi
\begin{document}

%%%%%%%%% TITLE
\title{TKD: Temporal Knowledge Distillation for Active Perception}

% Authors at the same institution
%\author{First Author \hspace{2cm} Second Author \\
% Authors at different institutions
\author{Mohammad Farhadi \\
Arizona State University\\
{\tt\small mfarhadi@asu.edu}
\and
Yezhou Yang \\
Arizona State University\\
{\tt\small yz.yang@asu.edu}
}

\maketitle
\ifwacvfinal\thispagestyle{empty}\fi

%%%%%%%%% ABSTRACT
\begin{abstract}
Deep neural network-based methods have been proved to achieve outstanding performance on object detection and classification tasks. Despite the significant performance improvement using the deep structures, they still require prohibitive runtime to process images and maintain the highest possible performance for real-time applications. Observing the phenomenon that human visual system (HVS) relies heavily on the temporal dependencies among frames from the visual input to conduct recognition efficiently, we propose a novel framework dubbed as TKD: temporal knowledge distillation. %which is a combination of light and heavy convolutional neural networks  or CNNs. 
This framework distills the temporal knowledge from a heavy neural network-based model over selected video frames (the perception of the moments) to a light-weight model. To enable the distillation, we put forward two novel procedures: 1) a Long-short Term Memory (LSTM)-based key frame selection method; and 2) a novel teacher-bounded loss design. To validate our approach, we conduct comprehensive empirical evaluations using different object detection methods over multiple datasets including Youtube-Objects and Hollywood scene dataset. 
%We first observe that our model requires much lower runtime while maintains high enough recognition accuracy.  
Our results show consistent improvement in accuracy-speed trad-offs for object detection over the frames of the dynamic scene, compared to other modern object recognition methods. It  can  maintain  the  desired accuracy with  the throughput  of around  $220$ images  per second. Implementation:  \url{https://github.com/mfarhadi/TKD-Cloud}.

\end{abstract}

%%%%%%%%% BODY TEXT
\section{Introduction}
\label{sec:intro}

Object detection plays a critical role in a variety of mobile robot tasks such as obstacle avoidance \cite{carrio2018drone, yaghoubi2019worst}, detection and tracking \cite{breuers2018detection} and object searching \cite{ye2018active,ye2019gaple}. During the last decade, we have witnessed the great success of Convolutional Neural Networks (CNNs)-based methods in the object detection task. This success has led researchers to explore deeper models such as RetinaNet \cite{lin2018focal} or Faster-RCNN \cite{ren2015faster}, which yield high recognition accuracy. The ``secret'' sauce behind the success of these deeper and deeper CNNs models is the stacking of repetitive layers and increasing the number of model parameters \cite{chen2017learning}. This practice becomes possible while the applications are running on infrastructures with high processing capabilities.

\begin{figure}[t!]
    \centering
    \includegraphics[trim=1cm 1cm 0cm 1cm, clip=true,width=7.5cm]{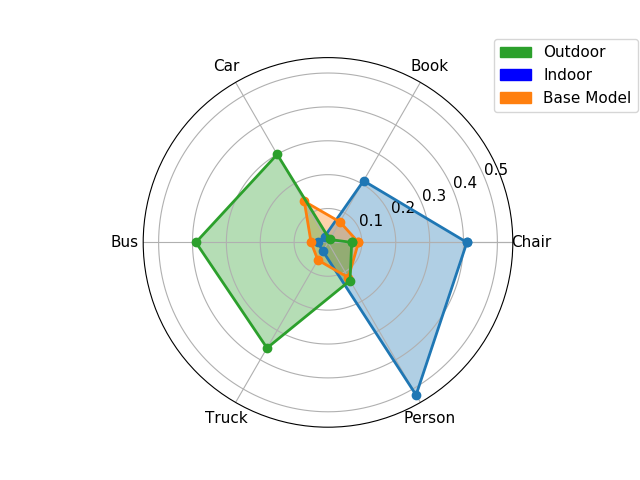}
    \caption{An illustration of our TKD model's actual performance: F-1 score distribution over example object categories in different environments using TKD.}
    \label{set-subset}
    \vspace{-15pt}
\end{figure}
 However, the disadvantages of this practice are obvious and the high performance is achieved by the significant growth of the model complexity:  stacking up layers and increasing the model parameters which are computationally expensive and also increase the inference time significantly.  Hence, these models are not suitable for real-time and embedded visual processing systems, and thus impede their deployment in the era of intelligent robots and autonomous vehicles.  The same concerns also lie in the energy conservation and computation limits, since deep models require a large number of matrix multiplications, which are time-consuming and energy-demanding for mobile applications. 

The aforementioned concerns trigger various approaches, such as using the alignment of memory and SIMD (Single Instruction, Multiple Data) operations to boost matrix operations \cite{gong2014compressing}. %{specific hardware (FPGA) solutions\cite{qiu2016going}}, and network compression methods  \cite{han2015deep,kim2015compression,zhang2016accelerating}.
More recently, studies \cite{chen2017learning} and \cite{hinton2015distilling} proposed transferring the knowledge of deep models to shallow models while maintaining the recognition accuracy.
%However, a significant performance gap still exists between these model and the state-of-the-art models such as Yolo-v3\cite{redmon2018yolov3}, ?? and ??.
Although these approaches do improve the model efficiency, they ignore the temporal dependencies among the frames from dynamic scenes, which is one of the critical capabilities to maintain high recognition accuracy while being energy-aware. %In other words, they reduce the computation overload by trading a large chunk of recognition accuracy off from the original model.
%, specially in the object recognition tasks.

%There is a rich structure in object detection tasks which is neglected; in real world images are highly structured over both the spatial and temporal dimensions. For instance, one would expect a smart system would consider recognition mainly from a small subset of objects that are relevant to a certain scene (a.k.a, witnessing a bed in the kitchen is very unlikely). However, the most of object detection methods consider all possible objects ignoring the context information from the scene.  {\color{red} Yezhou: the idea of temproal knowledge distillation is not described in the introduction yet. Integrate the Motivation part back with the introduction part helps maintain the follow.}

The motivation for our TKD model comes from the visual adaptation phenomenon observed in the Human Visual Systems (HVS). Visual adaption involves temporary changes in the human perception system when exposed to intense or new stimulus and by the lingering aftereffects when the stimulus is removed \cite{webster2015visual}. Other studies from \cite{webster2015visual} show that the visual system adapts to the changes in the environment and this adjustment can happen in a few milliseconds. More specifically, a study from \cite{clifford2007visual} reveals that the face recognition process happens at a higher level of cognition, and later at the stage of visual encoding, we observe that the sensory systems adapt itself to the prevailing environment. This shows that HVS relies heavily on the prior estimation of the objects' appearance distribution to improve the perception capability at the current time-stamp.    

Moreover, the adaptation happens both in the ``low'' and ``high'' level visual features. The human visual system adapts to the distribution of ``low-level'' visual features such as color, motion, and texture, as well as the ``high-level'' visual features such as face classification including identity, gender, expression, or ethnicity \cite{webster2015visual}. This adaptation can be both short-term and long-term. For instance, our perception system adapts itself to the general visual features of the environment which we are living in for a long time such as faces and colors (like training a model). Also, it can adapt itself dynamically when the environment changes,  for example, moving from the indoor environment to the outdoor \cite{webster2015visual} (like adapting a shallow model). This adaptation capability is essential for our HVS to perform recognition well and efficient,  with low energy consumption.

%These findings motivate us to design the Temporal knowledge distillation model (TKD), a visual recognition computing model which yields different behaviors based on the distribution shift in space. This model can adapt itself to  the changes in the environment and reach to higher accuracy.  

Inspired by the aforementioned findings, we design our TKD framework that utilizes the knowledge distillation techniques. It transfers temporal knowledge from the heavy model to a light model to boost visual processing efficiency while maintaining the heavy model's (a.k.a., oracle model) performance.  Figure \ref{set-subset} illustrates the overall goal of this work. In this figure, we show how TKD improves recognition accuracy over different scenes, compared to the oracle model which we assume to be a perfect model. Also, we show the baseline model which is a tiny model with low accuracy compared to oracle recognition due to a much lower number of parameters. TKD achieves higher accuracy by adapting itself to the observed environment. In the case of an indoor scene, the TKD recognition accuracy improves significantly over objects which are more probable to be observed inside a building. In the outdoor case, TKD recognition accuracy improves over the objects such as a car, bus, and truck which are more probable to be observed outside. For a similar amount of model parameters as the baseline tiny model, the TKD will achieve much better performance over the more probable objects by dynamically learning from the oracle model. 

%{\color{red} Yezhou: one more paragraph needed to summarize our contributions. }
To summarize our contributions: 1) we propose an end-to-end trainable framework to transfer the temporal knowledge (a.k.a., the perception of the moment) of the oracle model to the student model; 2) we propose a novel teacher-bonded loss for knowledge distillation which has a simple structure and performs inferences briskly; and 3) we propose an efficient method to select key frames from the dynamic scene, that indicate the right timing to train student model and to improve the detection accuracy. We design and conduct empirical experiments on both the public datasets (the Youtube Object dataset and the Hollywood Scene dataset) as well as on two long videos with multiple scene changes, which validate each of the aforementioned novel design choices, by observing a fast object recognition performance while maintaining high detection accuracy.       

\section{Related Work}

%{\color{Red} First paragraph need to be updated due to similarity with HPEC}

Visual recognition systems, ranging from object recognition~\cite{lin2018focal}, action recognition~\cite{lea2016learning}, to  scene recognition~\cite{zhou2014learning} have gained attention in recent years. Significant improvements in recognition accuracy have resulted in economic and societal benefits in AI applications such as autonomous vehicles \cite{khayatian2019crossroads+,khayatian2018rim}, and IoT systems \cite{tonekaboni2018edge, tonekaboni2018scouts}. 

\noindent\textbf{Object Detection:} Object detection methods based on Convolutional Neural Networks (CNNs) have shown promising results over the past years. There are two main types of object recognition systems which are based on CNNs, one-stage, and two-stage. In one-stage methods, we classify and localize objects in one-stage. Images, when forwarded through the network produce a single output which is then used to classify or localize objects. Some examples of one-stage methods are Yolo~\cite{redmon2018yolov3}, RetinaNet~\cite{lin2018focal} and DSSD~\cite{fu2017dssd}. These models are faster compared to other methods due to ruining in a single stage. The second types of models are two-stage methods in which classification and localization happen as two different stages, using classification networks and region proposal networks respectively. Two famous two-stage models are FasterRCNN~\cite{ren2015faster}, R-FCN~\cite{dai2016r}. These models reach to higher performance with high intersection over union (IOU). However, Redmon et al. \cite{redmon2018yolov3} showed at lower IOU (IOU=0.5) one-stage models can perform the same accuracy as two-stage models.

\noindent\textbf{Model Compression:} Another thrust of work has focused on reducing the resources consumption of CNNs (due to expensive computation and memory usage) by compressing the network structures \cite{han2015learning,rastegari2016xnor}. Network pruning is one of well-studied approach which removes unnecessary connections from CNN model, to gain inference speedup \cite{wen2016learning,iandola2016squeezenet}. Quantizing~\cite{han2015deep,farhadi2019novel} and binarizing~\cite{rastegari2016xnor,bank2019polar} are two other methods that have been used to reduce network size and computation load. These methods improve performance at the hardware level by reducing the size of weights at the binary code level. However, the standard GPU implementation remains challenging for these methods to achieve runtime speedup \cite{han2015learning}. Also, the advantages of these methods over other one-stage methods without the fully connected layers (the network pruning target in \cite{han2015deep}) is not clear. 

%\textbf{Hierarchical Models:} The use of adaptive structures is a relatively newer approach which process image using hierarchical structures\cite{shen2017fast,zhou2017adaptive,teerapittayanon2016branchynet,bengio2015conditional}.  \cite{teerapittayanon2016branchynet} proposed branchy-net which based on the network confidence to control to further processing. Other methods also adopt this idea and improve the decision process\cite{bolukbasi2017adaptive,figurnov2017spatially}. These methods are successful in image classification tasks. However, directly applying these methods for object detection still  remains difficult due to the inherent non-uniformity among different parts of natural images.  %{\color{red}{MO: non uniformity in different segments of natural images.}}

\noindent\textbf{Domain Adaptation:} Object detection in the real world still needs to address challenges such as low image quality, large variance in the backgrounds, illumination variation, etc. These could lead to a significant domain shift between the training, validation and test data. Consequently, the field of domain adaptation has been widely studied in image classification \cite{tzeng2014deep,lu2017unsupervised} and object detection \cite{chen2018domain,dai2018dark} tasks. These methods improve accuracy on well-known bench-marking datasets. Nevertheless, they typically adopt an offline domain adaptation procedure and do not concern with domain-change during the inference stage. 

\noindent\textbf{Knowledge Distillation:} Knowledge distillation is another approach to boost accuracy in CNNs. Under the knowledge distillation setting, an ensemble of CNN models or a very deep model will serve as the teacher model, which transfers its knowledge to the student model (shallow model). Hinton et al.~\cite{hinton2015distilling} proposed a method to apply teacher prediction as a ``soft-label'' and distill teacher classifier's knowledge to the student. Moreover, they proposed a temperature cross entropy instead of $L2$ distance as the loss function. Romero et al.~\cite{romero2014fitnets} proposed a so-called ``hint'' procedure to guide the training of the student model. There are also other approaches to distill knowledge between different domains such as from RGB to depth images \cite{gupta2016cross,su2016adapting}.
Knowledge distillation has been also applied to the object detection task. Chen et al.~\cite{chen2017learning} proposed a method which adopts all of the soft labeling (labels generated by the teacher), the hard labeling (the ground truth) and the hint procedure to transfer knowledge from the teacher with deep feature extractor to the student with a shallow feature extractor. They adopt a two-stage method (FasterRCNN~\cite{ren2015faster}) in their system. Mehta et al.~\cite{mehta2018object} applied the same procedure to one stage method (Tiny-Yolo v2). 

Mullapudi et al.~\cite{mullapudi2018online} proposed an online model distillation for efficient segmentation. They adopt a light CNN model as a student and a heavy model as a teacher. At the inference time, the student model is trained periodically using the teacher knowledge. However, the naive usage of a fixed period may not be efficient in their approach. Moreover, their shallow model struggles to handle emerging new objects in the scene when these objects are observed in the middle of the fixed period.  
Here, ours is able to select the period length based on the incoming frames, by which TKD could trigger re-training and thus detecting the emerging new objects, as demonstrated experimentally in Sec.~\ref{sec:exp}. %this part needs clarification. 

\begin{figure*}[ht]
    %\vspace{5}
    \centering
    \includegraphics[trim=0.5cm 4.9cm 0.3cm 1.7cm, clip=true,width=16cm]{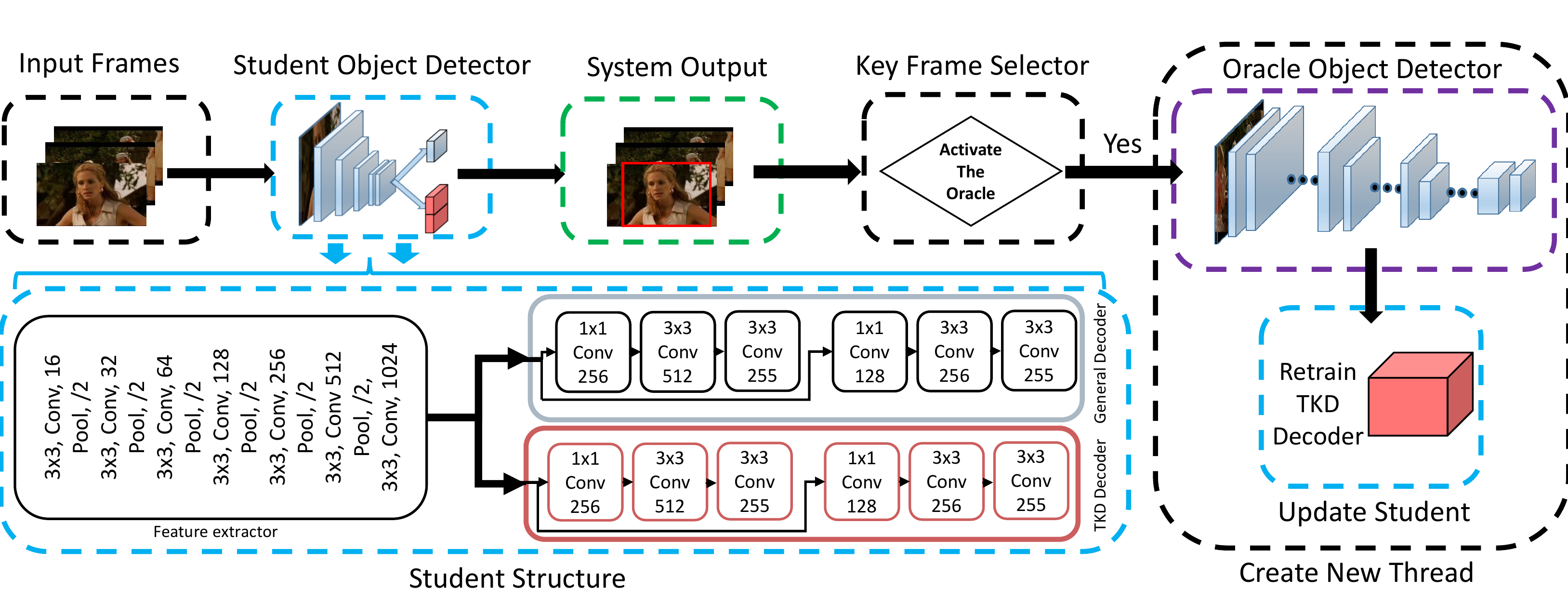}
    \caption{An overview of TKD (Temporal Knowledge Distillation): A low-cost student model is tasked to detect objects in the main thread. To retain high accuracy, a key frame selector decides to activate the oracle model and adapt the student over the environment. Since the execution of the Oracle model and retraining the student model occurs in separate threads, it does not have a significant effect on the inference latency.}
    \label{TKD_fig}
    \vspace{-11pt}
\end{figure*}

\section{Temporal Knowledge Distillation}
\label{distillation}
The conventional use of knowledge distillation has been proposed for training CNNs based classification  models. In these models, we have a dataset $(x_{i},y_{i}), i=1,2,...,n$ where $x_{i}$ and $y_{i}$ are input images and the class labels. The student model is trained to optimize the following general loss function (with $\beta$ is a modulation factor):
\begin{equation}
\small
\setlength{\abovedisplayskip}{5pt}
\setlength{\belowdisplayskip}{5pt}
\centering
\begin{aligned}
& O_s = Student(x); O_t= Teacher(x), \\
& L(O_s,(y,O_t))  = \beta L_{gt}(O_s,y) + (1-\beta)L_t(O_s,O_t), \label{eq:a}
\end{aligned}
\end{equation}
where $L_t$ is the loss using teacher output ($O_t$) and $L_{gt}$ is the loss using ground truth $y$ \cite{mehta2018object,chen2017learning,hinton2015distilling}.

In addition to the classification task, object detection also could benefit from the knowledge distillation procedure. However, it's not as straightforward as the classification task. Most notably, the teacher model's output may yield misleading guidance to the student model~\cite{chen2017learning}. The teacher regression result can be contradictory to the ground truth labels, also the output from the teacher regression module is unbounded. To address these issues, \cite{chen2017learning} proposed a procedure to only adopt teacher's output at beneficial times. For a one-stage object detection setting,  \cite{mehta2018object} optimized the student model with a similar loss function to Eq.~\ref{eq:a}.

In this paper, we propose a novel and bio-inspired way of adopting the teacher model's knowledge. Namely, temporally estimating the expectation of object labels, their sizes, and shapes based on the previously observed frames or $E[y_i| \alpha_1, 	\alpha_2, ... ,	\alpha_{i-1}]$ where $y_i$ is our objects label and $\alpha$ our observations. This expectation changes in time by camera or objects movements, and/or the changing of the field of view. Here, we utilize this extracted knowledge to improve object detection performance. Unlike the previous work such as \cite{mehta2018object,chen2017learning}, we are not aiming to improve the feature extractor and/or the general knowledge of the student model. We optimize the decoder inside the student model to adapt it to the current environment. It is done by increasing the likelihood of objects which are more frequently found from the previous observations. Since the model requires online training during the inference stage, it should be able to address the following challenges:

\begin{enumerate}[noitemsep,nolistsep]
\item Training is a time consuming procedure, running it at the inference stage hurts model efficiency;
\item Selecting the key frames accurately on which the student model needs to be adapted;
\item Objects with low appearance probability may not be detected by the student model after adaptation;
\item The oracle model still introduces noise at locations where there are no objects. Simply training the student model with noisy oracle output decreases the accuracy.
\end{enumerate}

In the following section, we will introduce our approach to address these challenges respectively.

\section{Our Approach}
\label{sec:TKD_structure}

In this work, we adopt Yolo-v3 (as teacher) and Tiny-Yolo v3 (as student) \cite{redmon2018yolov3} as the base object detection methods. These two models are  one-stage object detection models. In both models, object detection is conducted at various layers. The middle layers are used to detect large objects and the last layers to detect small objects. Studies \cite{redmon2018yolov3}, \cite{mullapudi2018online} and \cite{lin2018focal} showed that this  strategy successfully improves the object detection accuracy with a significant edge.

As mentioned in Section~\ref{sec:intro}, the overall objective of our system is to estimate the expectation of object labels, their sizes, and shapes on the temporal domain and to improve the performance of the student model. Following this intuition, we put forward a mechanism with a combination of an oracle model (which we consider it as the best possible model) and a student model (which is fast but has considerably lower accuracy compared to the oracle). We are transferring the temporal knowledge of the oracle model to the student model at the \underline{inference time}. By transferring this knowledge, the student model adapts itself to the current environment or scene. Without loss of generality, We select Yolo-v3 object detection model as the oracle model due to its reliable and dominating performance compared with other one-stage methods. We select Tiny-Yolo model \cite{redmon2018yolov3} as the student model due to its high base frame rate and having a similar model structure with the Yolo-v3.    

\subsection{The TKD Architecture}
%We learn strong and efficient objects detector by transferring temporal knowledge of a high capacity oracle to the student in time space. 
We show our overall framework in Figure \ref{TKD_fig}. In the student model, we include two decoders as the TKD decoder and the general decoder. Then, the pre-trained Yolo-v3~\cite{redmon2018yolov3} is adopted as the oracle. We run the Oracle model with the input image and the weights of student's TKD decoders get updates at specific frames  from the oracle model's result. Finally, we design a decision procedure using an LSTM model, to generate the signals that indicate the right timing to use the Oracle knowledge. 

Specifically, we train Tiny-Yolo with a general decoder over the COCO dataset~\cite{lin2014microsoft}. The design of Tiny-Yolo has two general decoders to improve the accuracy of different object sizes. We first make a copy of the general decoders bounded together as TKD decoder.  The TKD decoder is updated during the inference stage. We only update the last three layers of Tiny-Yolo and treat it as the decoder, since it yields enough performance in practice. %{ We observed that one of the most time consuming parts of Yolo is the detection layer which generate bounding boxes and labels from decoder output tensor. By considering this fact,}
We keep the general decoder from Tiny-Yolo together with the TKD decoder to make the final detection. TKD decoder and general decoder are executed in two parallel threads which do not increase the latency. This will preserve the chance of detecting viable objects  addressing the challenge (3) in Sec. \ref{distillation}.  

%regardless whether  from the previous observation. 

\subsection{Distillation Loss}

Before describing our distillation loss, we provide a brief overview of the other distillation loss functions. First, Chen et al. \cite{chen2017learning} proposed a combination of hint procedure and weighted loss function. They generate boxes and labels using both the student and the teacher model, then calculate two loss values comparing the teacher's output and the ground truth. In the end, they sum up the weighted loss values. If the student model outperforms the teacher model, they continue training only using ground-truth supervision. More recently, Mehta et al. \cite{mehta2018object} applied the similar procedure to the one-stage object detection models (Tiny-Yolo v2 with some modification). They generate bounding boxes and labels, and apply Non-Maximum Suppression (NMS) to these boxes and then follow the loss function to optimize the student model. The loss is defined in the following equation:
\begin{equation}
\small
\setlength{\abovedisplayskip}{5pt}
\setlength{\belowdisplayskip}{5pt}
\begin{aligned}
  L_{final}= L_{bb}^{C}(b_i^{gt}, \hat{b_i}, b_i^{T}, o_i^{T}) + L_{cl}^{C}(p_i^{gt}, \hat{p_i}, p_i^{T}, o_i^{T})\\
  + L_{obj}^{C}(o_i^{gt}, \hat{o_i}, o_i^{T}) ,
  \end{aligned}
  \label{eq:c}
\end{equation}
where $L_{bb}^{C},L_{cl}^{C},L_{obj}^{C}$ are objectness loss, classification loss and regression loss which are calculated using both ground truth and the teacher output. Also, $\hat{b_i},\hat{p_i},\hat{o_i}$ are bounding box coordinates, class probability and objectness of the the student model. $b_i^{gt},p_i^{gt},o_i^{gt}$ and $b_i^{T},p_i^{T},o_i^{T}$ are values derived from ground truth and the teacher model output. 

In our study of the Yolo-v3 and Tiny-Yolo models, we noticed that the detection layer is the most computationally expensive part. In this layer, several processes are done (sorting, applying softmax to classification cells, removing low confidence boxes, etc.) to produce bounding boxes and then applying NMS to these boxes. These processes are computationally slow due to the multiple steps of processing, and also running over CPU by the implementation. Consequently, directly adopting these loss functions will be also computationally expensive during the inference stage.

With this observation, we adopt the mean square error (MSE) between the tensors generated by the student decoder and the oracle decoder, which should be the fastest method. However, the side effects are also notorious. The oracle model generates noises over some parts of frame which have no object existences; hence directly forcing the student model to retrain will hurt its performance.

\begin{figure*}[t]

\centering
\begin{subfigure}{0.6\textwidth}
  \centering
  %\vspace{5}
  \includegraphics[trim=0.2cm 2.1cm 0.2cm 3.2cm, clip=true,width=10.4cm]{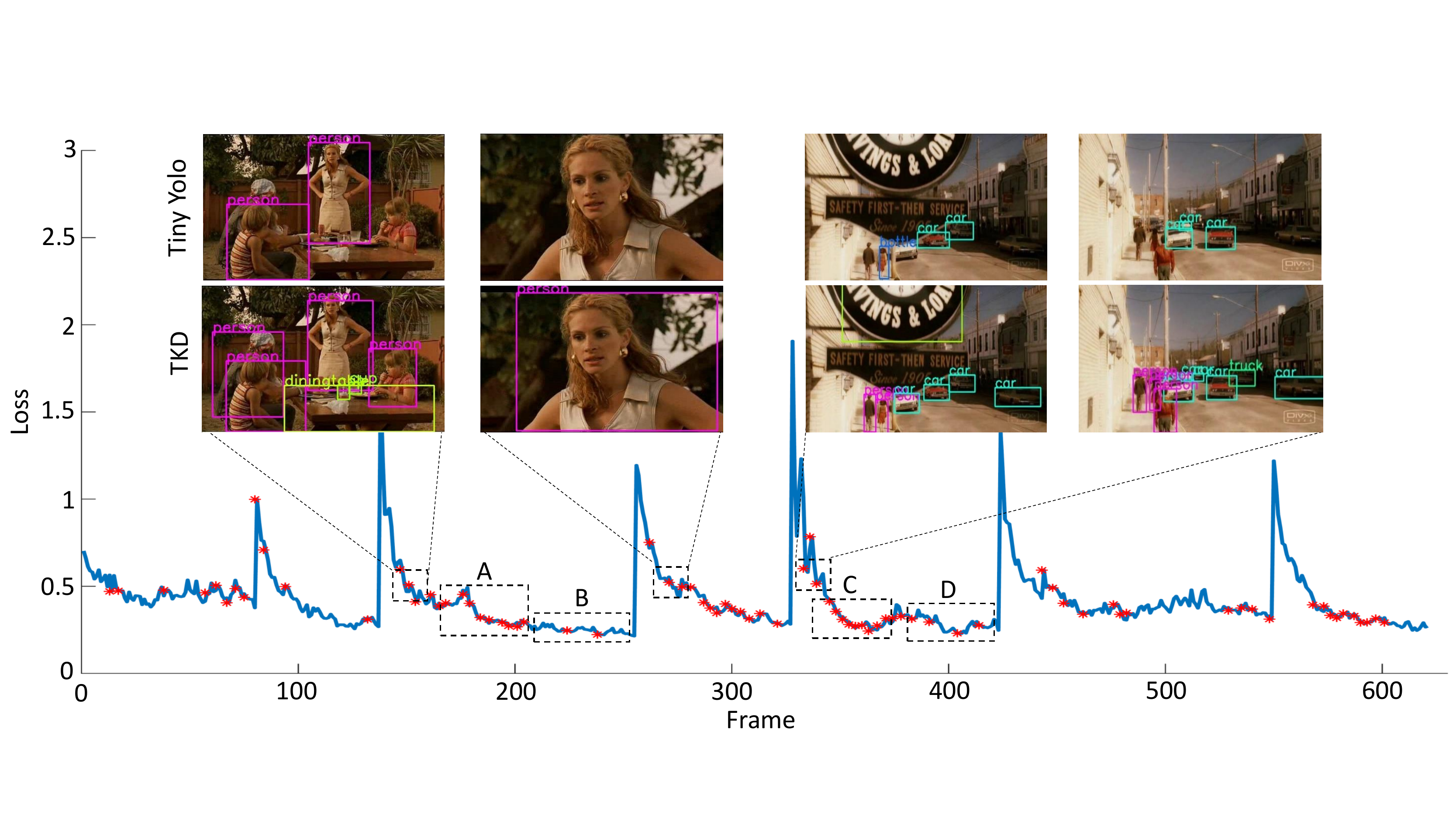}
  \caption{Key frames selected using TKD over two scenes from the Hollywood scene Dataset \cite{marszalek09}. The red crosses indicate the key frames selected by our method. See further discussion in Sec.~\ref{sec:discussion}.}
  \label{TKD_loss}
\end{subfigure}%
\begin{subfigure}{0.4\textwidth}
  \vspace{-13pt}
  \centering
  \includegraphics[trim=1.5cm 3cm 7.8cm 0.9cm, clip=true,width=6.9cm]{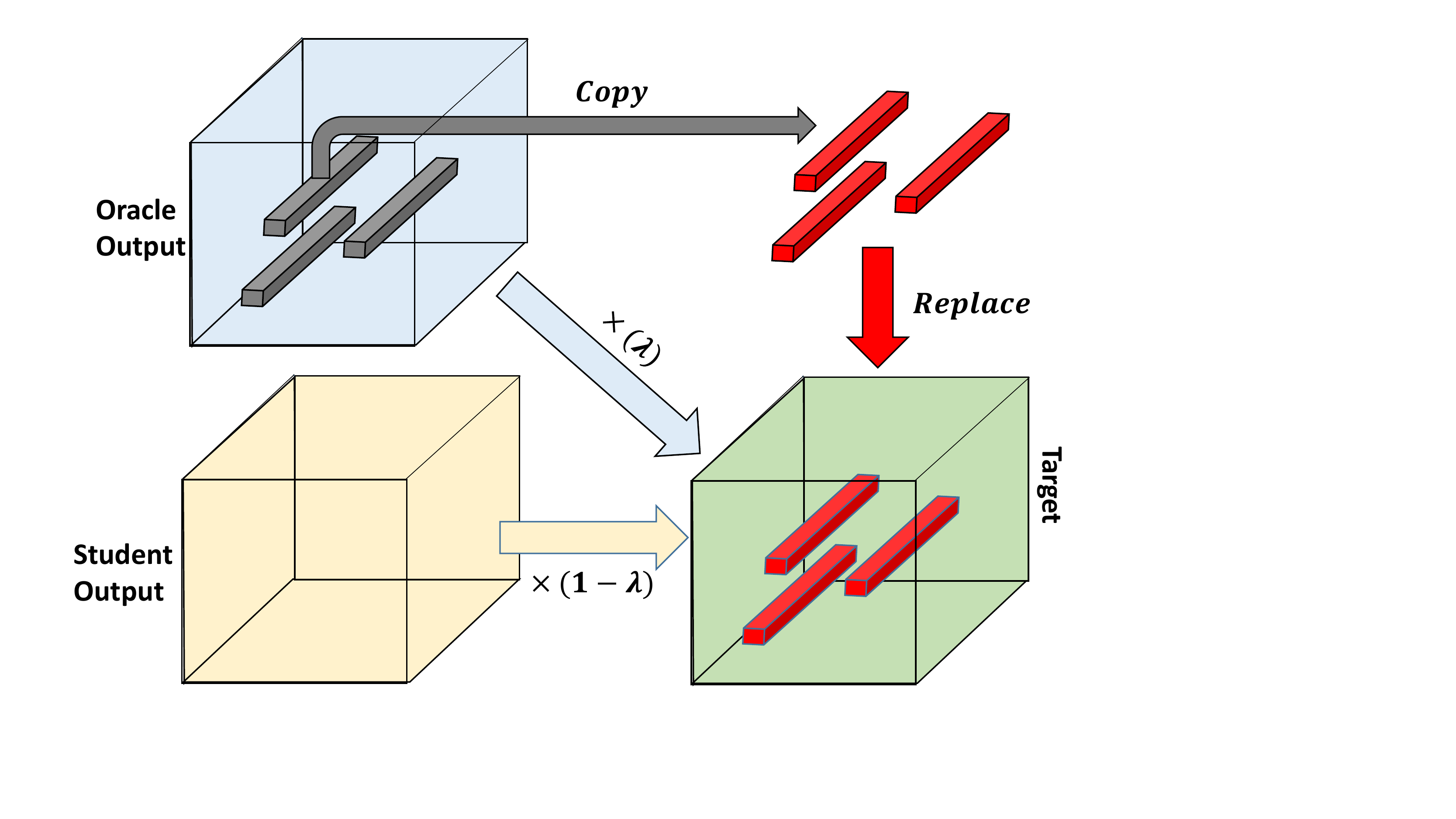}
    \caption{Target tensor composition.}
    \label{Tensor_copy}
\end{subfigure}
\vspace{-15pt}
\caption{TKD key frame selection and loss function.}
\vspace{-5pt}
\end{figure*}

%\begin{figure}[ht!]
%    \centering
%    \includegraphics[trim=1.5cm 3cm 7.8cm 0.9cm, %clip=true,width=8.0cm]{./images/Target_creating2.pdf}
%    \caption{Target tensor composition.}
%    \label{Tensor_copy}
%    \vspace{-10}
%\end{figure}

Another approach could be calculating the MSE between the tensor cells which have high confidence of object existence. But, the approach will hurt the student's recognition accuracy too. By applying this loss function, the student model tends to generate redundant detection boxes which yield a larger number of false positives.

To alleviate the downsides of both loss designs and still to preserve their advantages, we introduce a novel loss by a combination of them in Equation ~\ref{eq:b}: 
\begin{equation}
\small
\setlength{\abovedisplayskip}{5pt}
\setlength{\belowdisplayskip}{5pt}
\centering
\begin{aligned}
 & L_{final}= \sum \lVert T_s^H - T_o^H \rVert_2^2\\
 &   + \sum \lVert T_s^E - ((\lambda * T_s^E) + ((1-\lambda) * T_o^E)) \rVert_2^2,
  \end{aligned}
  \label{eq:b}
\end{equation}
where $T_s^H \& T_o^H$ are the student and oracle cells with a high chance of object existences and $T_s^E \& T_o^E$ are the cells with a low expectation. More specifically, the first part on the left side of Eq.~\ref{eq:b} calculates the MSE between the parts which have high confidence of objects. The second part calculates a modulated MSE between the cells with a low expectation from both the oracle output tensor and the student output tensor. Here, $\lambda$ is the modulation factor. Figure \ref{Tensor_copy} shows the procedure of creating the target tensor. 

 By using this loss function,  the student model will have a lower chance to generate extra false positives. Also, it would not strictly force the student model to mimic the oracle exactly.  We aim to partially address the challenges 1) and 4) in Sec. \ref{distillation}, with such a fast and effective loss function.  

%\begin{figure}[h!]
%    \centering
%    \includegraphics[trim=3.9cm 9.3cm 4.6cm 5.5cm, clip=true,width=8cm]{./images/flowchart.pdf}
%    \caption{Key frame selection procedure.}
%    \label{TKD_flowchart}
%\end{figure}
%\begin{figure*}[ht!]
%    \centering
%    \includegraphics[trim=0.0cm 2.1cm 0.2cm 2cm, clip=true,width=15cm]{./images/loss_changes4.pdf}
%    \caption{Key frames selected using TKD over two scenes from the Hollywood scene Dataset \cite{marszalek09}. The red crosses indicate the key frames selected by our method. See further discussion in Sec.~\ref{sec:discussion}.}
%    \label{TKD_loss}
%    \vspace{-5}
%\end{figure*}

\subsection{Key Frame Selection}

Another crucial module to enable TKD working properly is a procedure to demonically select the time instances to train the student model during the inference stage.  Specifically, TKD seeks the frames that by training over them the model has a higher expectation of reducing the loss, thus eventually improves the detection accuracy. For the rest of the paper, we denote these frames as the key frames. 

Selecting a larger number of frames as the key frames will hurt the performance since re-training is computationally expensive; While selecting too few number of frames will hurt the detection accuracy as the student may not align well with the oracle model in time. Thus, an effective and fast procedure to select the key frames is highly desired to yield a positive effect on the system's performance.

We propose a key frame selection procedure which is both efficient and also practical. %Figure \ref{TKD_flowchart} shows the decision flow  for the key frame selection module. 
First, we check the training prevention factor $\tau$. If the student model has been trained in any last $\tau$ frames; we will exit the key selection procedure. It is based on the reasonable assumption that if we have an environment change,  it typically takes $\tau$ frames that this change to be fully observable. Thus, when we train the student, training for the next $\tau$ frames would not be beneficiary. Second, we start our decision process which we formulate in Equations \ref{eq:d}: 
\begin{equation}
\small
\setlength{\abovedisplayskip}{5pt}
\setlength{\belowdisplayskip}{5pt}
\begin{aligned}
& I \in  \{0,1\} \left\{\begin{matrix}
 &0  &Do \ not \ distill \ knowledge, \\ 
 &1  &Distill \ knowledge, 
\end{matrix}\right.
\\
& I = LSTM(F_s) \ \vee \ I_R, \quad  I_R\sim B(2,P_t),
\\
& P_t=\left\{\begin{matrix}
 &max((P_{t-1}-0.05),0.05) \ \ \ \ \ \ \  \Delta L <\sigma,  &   \\ 
 &min(2P_{t-1},1.0) \ \ \ \ \ \ \ \ \ \ \ \ \ \ \ \ \ \ \ \ \  \Delta L >\sigma,   & 
\end{matrix}\right. 
\end{aligned}
\label{eq:d}
\end{equation}
where $I$ is the indicator that denotes our final decision. It takes the disjunction of the LSTM's output and the random module's output. We pass the features extracted from the student model $F_s$ (the last layer before the decoder) to the LSTM module (with one LSTM layer \& one fully connected layer) which outputs a signal indicating to train the student model or not.  Here, it is worth to note that we introduce another binary random module $I_R$ (with binomial distribution $B(2,p_t)$) which decides in a random fashion to train the student model or not. The random procedure is added as a safeguard in case the LSTM model outputs a sequence of erroneous decisions. %Function $I$ will activate training procedure if LSTM or random module become true. 
In the end, \uline{we update the LSTM module based on the result feeding back after the training procedure.} If the LSTM makes a correct decision where the observed loss decrease $\Delta L \ < \ \sigma$ wherein our experiments $\sigma=-0.1$, the random factor $P_t$ will be reduced by $0.05$. If the LSTM model makes a wrong decision, we update the LSTM model and double the random factor $P_t$. Figure \ref{TKD_loss} shows an example output of key frames selected by our method. We apply knowledge distillation selectively to a few numbers of frames which partially addresses the aforementioned challenges 1) and 2) in Sec. \ref{distillation}.
%In this section, we introduced the TKD object detection method. Temporal knowledge distillation in this method is fast and effective due to novel loss function, efficient key frame selection and light training procedure.

% \usepackage{multirow}
\begin{table*}[ht]
%\vspace{5}
\begin{center}
\scriptsize
\begin{tabular}{|c|c|c|c|c|c|c|c|c|c|c|c|c|}
\hline
\multicolumn{1}{|c|}{\multirow{3}{*}{\textbf{Method}}} & \multicolumn{6}{c|}{\textbf{Hollywood Scene Dataset}}                                                                                                                                                                        & \multicolumn{6}{c|}{\textbf{The pursuit of happiness}}                                                                \\ \cline{2-13} 
\multicolumn{1}{|c|}{}                                          & \multicolumn{2}{c|}{\textbf{IOU=0.5}}                                & \multicolumn{2}{c|}{\textbf{IOU=0.6}}                                & \multicolumn{2}{c|}{\textbf{IOU=0.75}}                               & \multicolumn{2}{c|}{\textbf{IOU=0.5}} & \multicolumn{2}{c|}{\textbf{IOU=0.6}} & \multicolumn{2}{c|}{\textbf{IOU=0.75}} \\ \cline{2-13} 
\multicolumn{1}{|c|}{}                                          & \multicolumn{1}{c|}{\textbf{AP}} & \multicolumn{1}{c|}{\textbf{F-1}} & \multicolumn{1}{c|}{\textbf{AP}} & \multicolumn{1}{c|}{\textbf{F-1}} & \multicolumn{1}{c|}{\textbf{AP}} & \multicolumn{1}{c|}{\textbf{F-1}} & \textbf{AP}       & \textbf{F-1}      & \textbf{AP}       & \textbf{F-1}      & \textbf{AP}       & \textbf{F-1}      \\ \hline
Random Selection                                                &0.71                                  &0.75                                   &0.54                                  &0.68                                   &0.48                                  &0.49                                   &0.65                   &0.65                   &0.55                   &0.58                   &0.35                   &0.43                   \\ \hline
Scene Change Detection                                          &0.68                                  &0.58                                   &0.47                                  &0.50                                   &0.23                                  &0.35                                   &0.54                   &0.58                   &0.45                   &0.53                   &0.35                   &0.44                   \\ \hline
Tiny-Yolo \cite{redmon2018yolov3}                                                      &0.45                                  &0.16                                   &0.38                                  &0.14                                   &0.10                                  &0.28                                   &0.37                  &0.11                   &0.25                   &0.10                   &0.08                   &0.06                   \\ \hline
Tiny-Yolo (73\%) + Yolo-v3 (27\%)                                  &0.60                                  &0.49                                   &0.59                                  &0.49                                   &0.44                                  &0.46                                   &0.58                   &0.47                   &0.52                   &0.46                   &0.39                   &0.44                   \\ \hline
TKD                                                             &\textbf{0.75}                                  &\textbf{0.76}                                   &\textbf{0.58}                                  &\textbf{0.69}                                   &\textbf{0.49}                                   &\textbf{0.50}                                    &\textbf{0.73}                   &\textbf{0.67}                   &\textbf{0.59}                   &\textbf{0.61}                   &\textbf{0.40}                    &\textbf{0.46}                   \\ \hline
\end{tabular}
\end{center}
\vspace{-15pt}
\caption{Performance of TKD with different training methods over Hollywood scene dataset and The pursuit of happiness.}
\label{TKD_hollywood}
\vspace{-12pt}
\end{table*}

\section{Experiments}
\label{sec:exp}
% \usepackage{multirow}

The presented theoretical framework suggests three hypotheses that deserve empirical tests: 1) TKD can perform visual recognition efficiently, without hurting the recognition performance significantly; 2) the novel loss function can improve online training of the decoder; 3) with our TKD frame selector mechanism, the overall system yields the best performance over other key-frame selection mechanisms, by locating the key frames more accurately (frames which training over them can improve TKD accuracy).     

To validate these three hypotheses, we evaluate TKD on the Hollywood scene dataset \cite{marszalek09}, YouTube-Objects dataset \cite{prest2012learning}, The Pursuit of Happyness~\cite{Muccino2008} and the office \cite{Daniels2013}. We have trained all the base models (RetinaNet \cite{lin2018focal}, FasterRCNN~\cite{ren2015faster}, Yolo-v3 and Tiny-Yolo \cite{redmon2018yolov3}) over MS COCO dataset \cite{lin2014microsoft}. We implemented the TKD as described in Sec.~\ref{sec:TKD_structure} with two different configurations. First, we perform the process of inference and distillation sequentially among the same thread; the other way, we perform the distillation in a separate thread and run the student and oracle in parallel, both architecture implemented using the PyTorch environment \cite{paszke2017automatic}.  All experiments are carried out on one single NVIDIA TITAN X Pascal graphics card. 
%We have made our implementations publicly available for the community to conduct further research.

\textbf{Hollywood scene dataset \cite{marszalek09}} has 10 classes of scenes distributed over 1152 video. In this dataset, videos are collected from 69 movies. The length of these video clips are from 5 seconds to 180 seconds. The length and diversity of video clips make this dataset a perfect candidate to evaluate our key selector method and the novel loss function.

\textbf{YouTube-Objects dataset \cite{prest2012learning}} is a weakly annotated dataset from YouTube videos, 10 object classes of the PASCAL VOC Challenge~\cite{everingham2010pascal} has been used in this dataset. It contains 9 and 24 video clips for each object class which length of these videos are between 30 seconds to 3 minutes. We used this dataset to evaluate TKD's overall performance due to its high-quality objects level annotations.   

\textbf{The pursuit of happyness \cite{Muccino2008} \& The office \cite{Daniels2013}} are two famous movie and TV series. These two video clips contain several scenes which have smooth transitions. The Pursuit of happyness serves a great testbed since it has scenes in different locations such as office, street, etc. It is also more close to the real world scenario from a camera of the intelligent agent. Also, the Office is selected as most of the scenes have been recorded in the same location which make it suitable for testing our novel loss function.     

\subsection{Ablation Study}
\label{sec:ablation}
As shown in table \ref{TKD_hollywood}, we compare different strategies to highlight the effectiveness of our proposed novel loss and key frame selector. We consider the output of the oracle model as ground truth and evaluating different methods over it. Here, we compare five methods: 1) TKD with random key frame selection; 2) TKD with Scene Change detection; 3) Tiny-Yolo without any training; 4) Combination of Tiny-Yolo and Yolo-v3 without training; 5) TKD with our proposed key frame selection method.

In the following experiments, we have set the $\lambda$ to be $0.4$ which is obtained heuristically. In \ref{sec:discussion}, we will go through the findings which we observed in our search for the best $\lambda$. 

\textbf{Random Selection:} Here, instead of selecting key frames by our proposed method, decision modules selects frames purely randomly for further processing. During the testing phase, the probability  is set to be $27\%$ (to make sure it selects more frames than our method ($25\%$ on average)). Random selection achieves $0.75$ $F_1$ score (IOU=$0.5$) in the Hollywood scene dataset and achieves $0.65$ $F_1$ score (IOU=0.5) in the pursuit of happiness. On average, it reaches a frame-rate of $89$ frames per second (FPS).

\textbf{Scene Change Detection:}This method uses the content-aware scene detection method \cite{castellano2018pyscenedetect}. It finds areas where the difference between two subsequent frames exceeds the threshold value and used them as key frames for training the student. We selected the threshold with the highest performance and accuracy to report. This method achieves $0.58$ $F_1$ score and $0.58$ $F_1$ score in the Hollywood scene dataset and The pursuit of happyness respectively. This method selected $24\%$ frames as key frames ultimately. On average, the system yields a $93$ FPS. 

\textbf{Tiny-Yolo without any training:} We test Tiny-Yolo \cite{redmon2018yolov3} to show the accuracy of a strong baseline model without temporal knowledge distillation. This model achieves $0.16$ $F_1$ score and $0.11$ $F_1$ score in the Hollywood scene dataset and The pursuit of happyness respectively,  which are significantly lower than the other mentioned methods. However, This model has $220$ FPS, the fastest among all.  

\textbf{Tiny-Yolo $+$ Yolo-v3 without training:} In this configuration, we used Tiny-Yolo and Yolo-v3 v3 \cite{redmon2018yolov3} together. We designed a random procedure which runs Yolo-v3 with a probability of $27\%$ and Tiny-Yolo for the rest of the times. This model achieves $0.49$ $F_1$ score and $0.47$ $F_1$ score in the Hollywood scene dataset and the pursuit of happyness respectively. Frame-rate approaches 89 FPS.  

\textbf{TKD with our key frame selection method:} Initially, we set $\tau$ (the training prevention factor) to $2$ (We observe that the transition between two scenes takes at least $2$ frames); along with setting the minimum random selection to $5\%$. In the Hollywood dataset, our method selects around $26\%$ of frames and the $F_1$ score achieves $0.76$ (IOU=0.5). In the pursuit of happyness movie, our method selects around $24\%$ of frames and the $F_1$ score reaches to $0.67$ (IOU=$0.5$). On average, the system achieves a frame-rate of $91$ FPS sequentially and $220$ FPS with running inference and knowledge distillation in parallel.

Table \ref{TKD_hollywood} lists the experimental results we observed with these variants. These experiments show, the TKD, while maintaining a similar frame-rate as other methods, it can achieve higher recognition accuracy. To further validate this claim, we conduct one additional experiment on a single-shot movie \cite{fish2013}, TKD selects $21\%$ and random procedure selects $27\%$ of the total frames for re-training. They reach comparable F1-score (TKD:$0.807$, Random:$0.812$), but our TKD
method uses 10400 frames less than the random one.

\begin{table}[t]
%\vspace{5}
\begin{center}
\scriptsize
\begin{tabular}{ccc}
\hline
\multirow{2}{*}{Method}                              & \multicolumn{2}{c}{IOU=0.5}                                             \\ \cline{2-3} 
                                                     & \multicolumn{1}{c|}{mAP}           & F-1 score                          \\ \hline
\multicolumn{1}{|c|}{RetinaNet-50~\cite{lin2018focal}}                   & \multicolumn{1}{c|}{0.45}          & \multicolumn{1}{c|}{0.44}          \\ \hline
\multicolumn{1}{|c|}{FasterRCNN~\cite{ren2015faster}}                & \multicolumn{1}{c|}{0.52}          & \multicolumn{1}{c|}{0.50}          \\ \hline
\multicolumn{1}{|c|}{Tiny-Yolo \cite{redmon2018yolov3}}                   & \multicolumn{1}{c|}{0.38}          & \multicolumn{1}{c|}{0.33}          \\ \hline
\multicolumn{1}{|c|}{Tiny-Yolo (73\%) + Yolo-v3 (27\%)} & \multicolumn{1}{c|}{0.44}          & \multicolumn{1}{c|}{0.45}          \\ \hline
\multicolumn{1}{|c|}{TKD}                            & \multicolumn{1}{c|}{\textbf{0.56}} & \multicolumn{1}{c|}{\textbf{0.55}} \\ \hline
\multicolumn{3}{c}{\textbf{Oracle (Teacher)}}                                                                                  \\ \hline
\multicolumn{1}{|c|}{Yolo-v3~\cite{redmon2018yolov3}}                        & \multicolumn{1}{c|}{0.60}          & \multicolumn{1}{c|}{0.62}          \\ \hline
\end{tabular}
%\label{TB2}
\end{center}
\vspace{-15pt}
\caption{Compression of accuracy (IoU=0.5) over Youtube object dataset.}
\vspace{-12pt}
\end{table}

\subsection{Overall Performance}

Table 2 shows mean average precision (mAP) and $F_1$ score for five different object detection models as well as our TKD method over the Youtube object dataset \cite{prest2012learning}. For the student models without oracle’s supervision, we train them to the best performance we could achieve. Not surprisingly, larger or deeper models with larger numbers of parameters perform better than shallower models, while smaller models run faster than larger ones. However, TKD achieves a high detection accuracy compare to RetinaNet, FasterRCNN, Tiny-Yolo, the combination of Tiny-Yolo and Yolo-v3 (same configuration which is described in Sec.~\ref{sec:ablation}). TKD's detection performance also approaches the performance of the oracle model (Yolo-v3). In this experiment, $25\%$ of frames have been selected for training using the proposed key frames selection method.  

\begin{figure}[t]
    \centering
    \includegraphics[trim=3.3cm 7.1cm 2.6cm 14.9cm, clip=true,width=7.8cm]{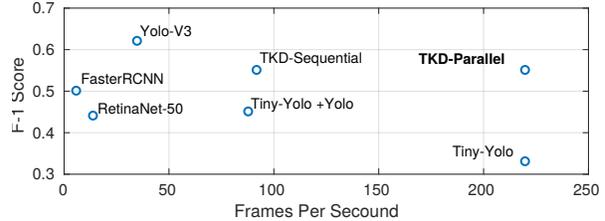}
    \vspace{-8pt}
    \caption{Accuracy and speed in Youtube-Objects Dataset.}
    
    \label{TKD_plot}
    \vspace{-5pt}
\end{figure}

To illustrate the accuracy-speed trade-off, we further plot them in Figure \ref{TKD_plot}, where we can see that the TKD archives higher accuracy compare to other shallow methods while still operating far above the real-time speeds with a $91$ FPS. The oracle model has a better detection accuracy, but it runs much slower than the TKD.  

%\vspace{-5pt}
\subsection{Further Study and Discussions}
\label{sec:discussion}
In this section, we provide further insight into the loss function design, the general knowledge distillation idea, and suggest an application of the proposed method.

\textbf{Loss function:} we studied the $\lambda$ effect over the number of true positives and false positives generated by TKD. All tests are done over an episode from The office \cite{Daniels2013}. We choose this video since it was recorded in one indoor environment, with a consistent objects distribution. Table \ref{loss_tune} shows the student model's detection accuracy varies with the different choices of $\lambda$.  At $\lambda=0$, we observed a lower number of false positives since a fewer number of frames ($5\%$) selected by the key frame selection module. With a low $\lambda$ (except at $0$), we observe an increase in false positives as the model tries to generate more boxes and loss function doesn't punish hardly enough onto the student model for generating false positives. With a high $\lambda$, we observe drops in the true positive rates since we are forcing the student to learn noises which are likely  introduced by the oracle model. Consequently, $0.4$ is empirically the best choice here, and we set it as the $\lambda$ value for all the experiments. 

\begin{figure}[t]
\small
    \centering
    \includegraphics[trim=3.9cm 10.5cm 4.68cm 11.8cm,clip=true, width=7cm]{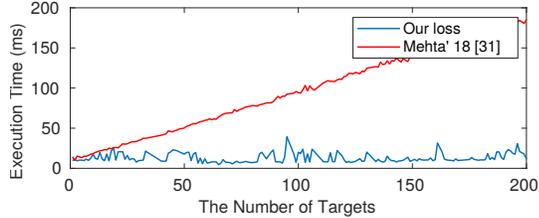}
    \vspace{-10pt}
    \caption{ Computational costs for loss functions.}
    \label{LossPerf}
    \vspace{-5pt}
\end{figure}

To validate our loss design, we further compare its performance with the one from Mehta et al. \cite{mehta2018object}, where the proposed loss is based on Non-Maximum Suppression algorithm. It is computationally more expensive in comparison with our approach. Figure~\ref{LossPerf} depicts that, an increasing number of targets from each frame will result in the increasing of execution time for calculating the loss function in \cite{mehta2018object}. Our loss design has an almost constant execution time, while the proposed loss function by \cite{mehta2018object} is linearly growing. 

\begin{figure}[t]
    \centering
    \includegraphics[trim=4.2cm 10.7cm 5.3cm 12.3cm,clip=true, width=7cm]{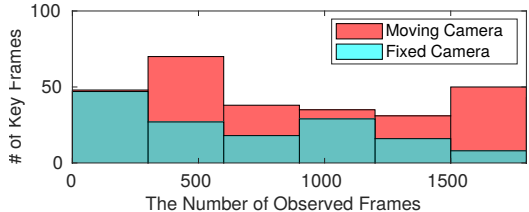}
    \vspace{-5pt}
    \caption{Key frames histogram.}
    \label{histo}
    \vspace{-5pt}
\end{figure}

%At $\lambda=0$, we observed lower number of key frames selection. We changed the key frame selection to random selection and scene change detection. The results shows that number of false positive passed the $10000$. Consequently, $F_1$ score decreases significantly. By a close look at the number of key frame selected by our proposed method, we observed that our method select less number of frames at low $\lambda$ which prevents drop in student precision accuracy. This shows our selection method have control over the overall performance of student model. 

% \usepackage{multirow}
%\vspace{-5}
\begin{table}[t]
%\vspace{5}
\begin{center}
\scriptsize
\begin{tabular}{|c|c|c|c|c|c|c|c|}
\hline
\multicolumn{2}{|c|}{ssd}                                                                  & \textbf{0} & \textbf{0.2} & \textbf{0.4}   & \textbf{0.6}  & \textbf{0.8} & \textbf{1} \\ \hline
\multirow{2}{*}{\textbf{\begin{tabular}[c]{@{}c@{}}IOU\\ 0.5\end{tabular}}} & \textbf{AP}  & 0.47       & 0.72         & 0.82           & \textbf{0.83} & 0.79         & 0.8        \\ \cline{2-8} 
                                                                            & \textbf{F-1} & 0.36       & 0.649        & \textbf{0.676} & 0.656         & 0.634        & 0.643      \\ \hline
\multicolumn{2}{|c|}{\textbf{\#TP}}                                                        & 3353       & 8570         & 8371           & 7806          & 7274         & 7438       \\ \hline
\multicolumn{2}{|c|}{\textbf{\#FP}}                                                        & 215        & 2952         & 1522           & 1129          & 814          & 841        \\ \hline
\end{tabular}
\end{center}
\vspace{-10pt}
\caption{Parameter study of $\lambda$ over the TKD.}
\label{loss_tune}
\vspace{-15pt}
\end{table}

\textbf{Temporal knowledge distillation:} Here, we take a closer look at the key selection module. Figure \ref{TKD_loss} shows its performance over two video clips from the Hollywood scene dataset. Red crosses are frames selected by our proposed method as key frames. At peaks, we have a scene change and logically these points would be the best candidate for training. Following this insight, we observe our model has a lag on detecting these points. Here, we argue that training over these frames is not the best one for improving the student model's accuracy. The scene detection method can identify these points yet table \ref{TKD_hollywood} shows it achieves lower  accuracy. Figure \ref{TKD_loss} shows the TKD after detecting a change in loss start stabilizing the model by selecting most of the frames (parts A \& C) and for the rest select less number of frames (parts B \& D).

The proposed key frame selection method leads to improved performance comparing with \cite{mullapudi2018online}'s. Figure~\ref{histo} shows that the number of selected key frames is adjusted based on the domain change. With the fixed camera case in which the domain does not change, the number of selected frames decreases along observing more frames (validated over the UCF Crime dataset~\cite{sultani2018real}). Indeed, for the case of a moving camera, more key frames are selected to adjust the TKD to the specific domain. Here, the method presented in \cite{mullapudi2018online} relies on a static strategy of selecting frames which are chosen manually at the beginning.        

%Our studies over key frame selection shows that, it's still far from the ideal selection by selecting unnecessary frames as key frames (we reach to better performance by handpicking the frames) though it yields the best empirical performance over other methods.           

For further evaluation, we applied TKD on one episode of the office TV series. Then, we test the trained student model over another episode without any re-training at the inference time. We observed an increase of precision by $6\%$ comparing to the case in which we use the original student model without applying TKD. The result demonstrates the domain adaption capability of our method. Furthermore, it maintains a high recall over other domains which indicates that unseen objects have a chance to be detected. With the method presented in \cite{mullapudi2018online}, the model loses its generality over unseen objects due to the practice of optimizing the overall model with the new frames.

%\textbf{Applications:} 
%One of the direct applications for TKD is to accommodate devices that are running on computing resources with limited memory size, such as on FPGA (field-programmable gate array). FPGAs are configurable integrated circuits which are more energy efficient compare to GPUs \cite{nurvitadhi2017can}. However, FPGA's low memory is a challenge for most of deep neural networks based method to reside. By using TKD, we are able to run the student model over FPGA and at the same time having the oracle model running on edge computing or GPUs. We have implemented a platform  for running light CNNs over FPGAs which TKD is the primary solution on this platform. We will open-source the platform upon the publication of this draft to facilitate further research.     

%\vspace{-5pt}
\section{Conclusion and Future Work}
In this paper, we propose a novel approach to distill temporal knowledge of an accurate but slow object detection model to a tinier model yielding a light and accurate object detection paradigm for robotic applications, called TKD. We conducted experiments on the Hollywood scene dataset, Youtube object dataset, the pursuit of happyness movie and the office TV series, and empirically validate that TKD maintains a high inference efficiency while achieving a high recognition accuracy. The accuracy even approaches the original oracle model for the object detection task.  

%{\color{red} MO: DONE. YZ: let me try.}

The promising experimental results we observed suggest several potential lines of future work: 1) the frame selection procedure could be further optimized to  be more selective while maintaining the recognition accuracy; 2) we plan to test our TKD model with an oracle model that follows the two-stage object detection manner; 3) TKD performance can future improve by adopting temporal features in video.

\noindent\textbf{Acknowledgment:} The National Science Foundation under the Robust Intelligence Program (\#1750082), and the IoT Innovation (I-square) fund provided by ASU Fulton Schools of Engineering, GPU and FPGA donations from NVIDIA and Xilinx,  are gratefully acknowledged.

{\small
\bibliographystyle{ieee}
\bibliography{egbib}
}

\end{document}